\documentclass[letterpaper, 10 pt, conference]{ieeeconf}  

\usepackage{amsmath,amsfonts}
\usepackage{algorithmic}
\usepackage{algorithm}
\usepackage{array}
\usepackage[caption=false,font=normalsize,labelfont=sf,textfont=sf]{subfig}
\usepackage{textcomp}
\usepackage{stfloats}
\usepackage{url}
\usepackage{verbatim}
\usepackage{graphicx}
\usepackage{cite}
\usepackage{hyperref} 
\usepackage{booktabs}
\usepackage{pifont}
\usepackage{booktabs}
\usepackage{tabularx}
\usepackage{makecell}
\usepackage[table]{xcolor}
\usepackage{threeparttable}
\usepackage{tikz}
\usetikzlibrary{arrows.meta,backgrounds,calc,fit,positioning}
\usepackage{float}
\usepackage{placeins}

\newcommand{\yes}{\ding{51}}
\newcommand{\no}{\textemdash}
\newcommand{\partialyes}{\textsc{Part.}}
\title{\LARGE \bf
\textit{GzDRL}: Reproducible and Scalable Deep Reinforcement Learning with Gazebo
}

\author{Amal Dev Haridevan, Junjie Kang, and Jinjun Shan
}

\begin{document}

\maketitle
\thispagestyle{empty}
\pagestyle{empty}

\begin{abstract}
We present \textit{GzDRL}, a novel single-process reinforcement learning (RL) framework for Gazebo that overcomes longstanding bottlenecks in scalable, reproducible robotics experimentation. Unlike conventional middleware-based RL-Gazebo integrations that suffer from nondeterminism and irreproducibility, \textit{GzDRL} introduces a systematic, middleware-free environment-stepping mechanism that directly synchronizes agent actions and physics updates. This design enables deterministic, high-throughput data collection, efficient vectorization, and reproducible RL training and evaluation. Comprehensive benchmarks demonstrate that \textit{GzDRL} achieves the highest workstation throughput among the evaluated frameworks while remaining competitive with GPU-accelerated simulators on laptop hardware, and maintains precise agent-environment synchronization, multi-agent scalability, and experiment-level reproducibility. We further validate sim-to-real transfer by deploying learned policies directly onto a physical quadrotor, without fine-tuning. Our results establish \textit{GzDRL} as an accessible and reproducible platform for advancing RL in robotics and automation. 
\newline
\newline
\indent \textbf{\emph{GitHub:}} \url{https://github.com/amaldevh/gz-drl}

\end{abstract}

\section{Introduction}
\begingroup
\renewcommand{\thefootnote}{}
\footnotetext{\scriptsize{This work has been submitted to the IEEE for possible
publication. Copyright may be transferred without notice, after which
this version may no longer be accessible.}}
\endgroup
Deep reinforcement learning (DRL) has achieved remarkable advances in aggressive multirotor flight and autonomous navigation \cite{kaufmann2023champion,song2021autonomous}. However, training such systems demands a vast number of simulated transitions, often necessitating high-throughput, scalable simulation platforms. While GPU-based simulators like Isaac Gym \cite{makoviychuk2021isaac} deliver impressive parallelism, Gazebo \cite{koenig2004design, gazebo_architecture_docs} remains the standard for robotics due to its faithful robot models, modular physics engines, broad sensor support, and seamless ROS \cite{macenski2022robot} integration.

Despite Gazebo’s strengths, most RL integrations decouple policy and simulation, exchanging actions and observations via ROS or Gazebo Transport~\cite{lucchi2020robo,martini2023pic4rl,haridevan2024ros2}. Middleware serialization and scheduling can result in actions being applied to outdated observations, disrupting the intended action and observation pairing. This introduces non-Markovian effects for standard feedforward policies, undermining learning stability, sample efficiency ~\cite{ramstedt2019real,anokhin2025handling}, and overall reproducibility. Moreover, middleware communication constrains parallel throughput. Addressing this bottleneck with a direct, single-process interface enables deterministic, high-throughput data collection while preserving compatibility with ROS-based software-in-the-loop (SITL) workflows. 

Prior work such as Gym-Ignition~\cite{ferigo2020gym} addresses middleware-related delays by embedding Gazebo directly within the RL environment. However, scalable batched execution—and its impact on environment step reproducibility and throughput—remains insufficiently explored.

\begin{figure*}[t]
\centering
\includegraphics[width=0.95\textwidth, trim={0 0 0 0},clip]{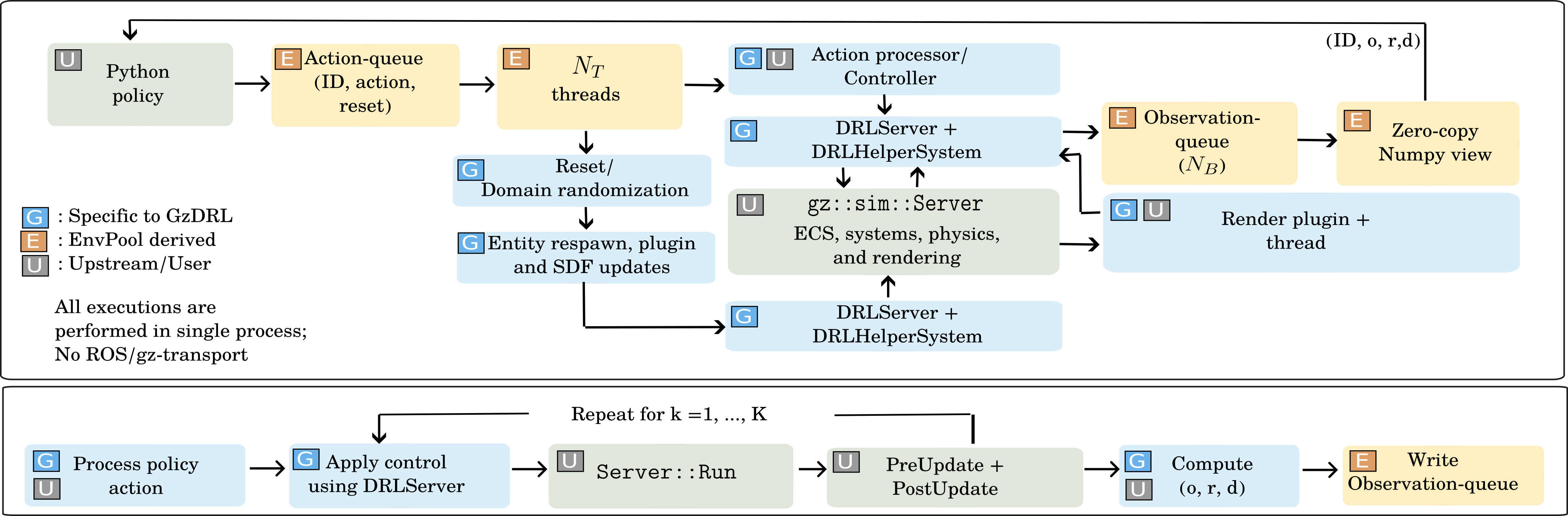}
\caption{\textit{GzDRL} system architecture (top).
An EnvPool-derived scheduler dispatches environment-indexed actions to
$N_T$ threads, which write an observation block after $N_B$ completions. Each environment contains a \textit{DRLServer} to interface with Gazebo for performing resets and executing actions. Internal execution pipeline from action processing to observation generation (bottom). The processing scripts can be provided by the user.}
\label{fig:gzdrl_architecture}
\end{figure*}

Recent efforts such as EnvPool \cite{weng2022envpool} have demonstrated the benefits of C++-based batched vectorization for scalable RL data collection. Yet, Gazebo’s programmatic server remains underutilized in DRL. In this work, we introduce \textit{GzDRL}: a novel, systematic environment-stepping mechanism that unifies direct entity-component-system (ECS) \cite{gazebo_architecture_docs}-based commands, controlled physics stepping, and efficient, zero-copy observation exposure. This architecture preserves deterministic agent-environment synchronization, enables scalable vectorization, and bridges Gazebo’s high-fidelity simulation with RL experimentation. Figure~\ref{fig:gzdrl_architecture} summarizes the proposed framework.

\begin{figure}[t]
\centering
\includegraphics[width=0.92\columnwidth]{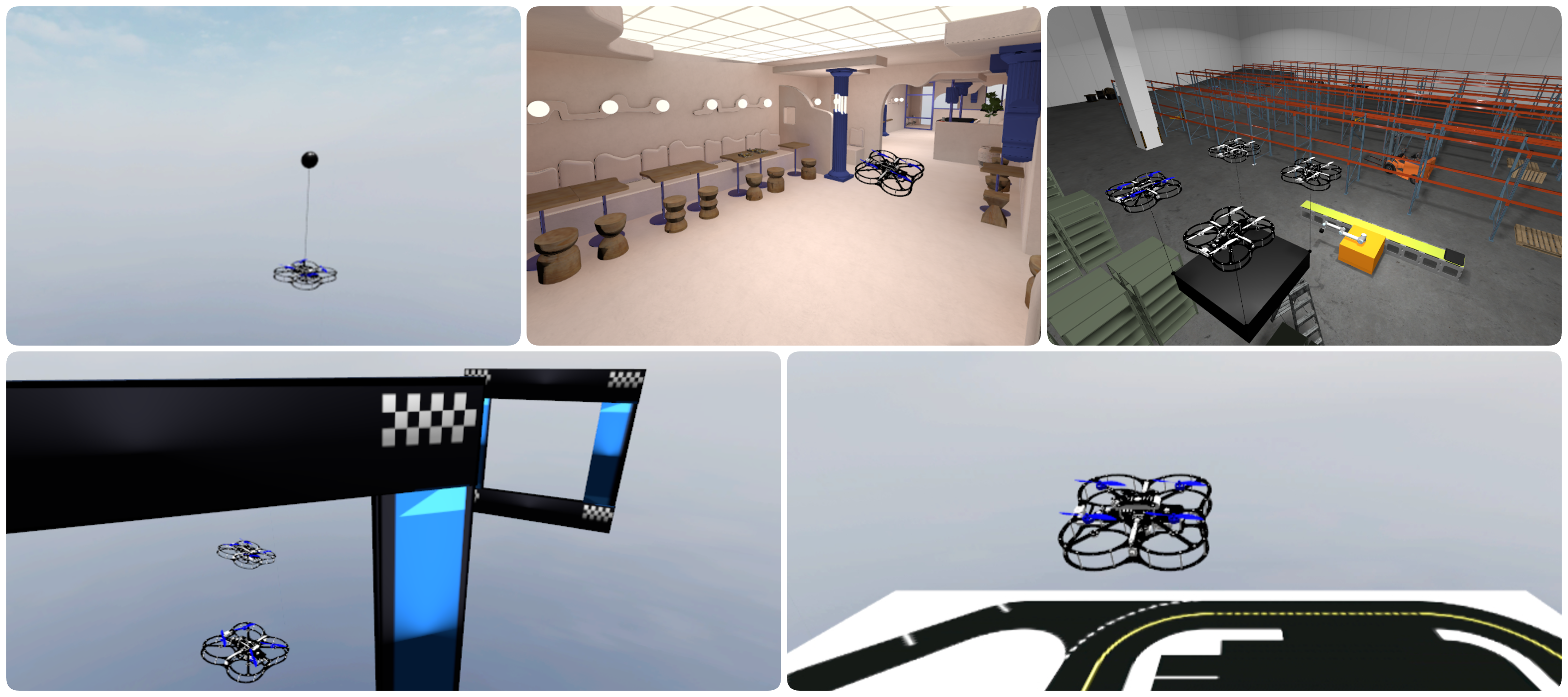}
\caption{Representative \textit{GzDRL} scenarios, ranging from drone racing to cooperative
payload transport.}
\label{fig:gzdrl_scenarios}
\end{figure}

This paper makes the following key contributions:
\begin{itemize}
\item Proposes a novel, systematic environment-stepping mechanism for reinforcement learning in Gazebo, enabling direct, middleware-free synchronization between agent and environment via ECS commands and controlled physics stepping.
\item Adapts EnvPool’s batched vectorization for Gazebo, supporting efficient multi-instance simulation with partial batches and zero-copy C++/NumPy \cite{harris2020array} observation sharing.
 \item A comprehensive experimental validation of the proposed
architecture, including state-transition and training reproducibility,
single- and multi-agent scalability on two hardware platforms,
control abstractions, runtime model randomization, and zero-shot
physical deployment, with CPU throughput reaching
$78.6\times10^3$ environment steps/s.
\end{itemize}
Figure~\ref{fig:gzdrl_scenarios} illustrates representative tasks, including single-robot flight and cooperative payload transport.

\section{Related Work}
Simulation platforms for robotics RL span diverse trade-offs between physical fidelity, scalability, and integration with real-world systems. RotorS \cite{furrer2016rotors} pioneered detailed multirotor dynamics in Gazebo with ROS integration. However, RotorS is limited in scalability and multi-agent extensibility due to reliance on middleware. Gym-Ignition~\cite{ferigo2020gym} embeds Ignition Gazebo within the RL process and enables parallel execution via independent environments, but does not present efficient batched execution or quantitatively evaluate the reproducibility of state transitions and RL training. In contrast, \textit{GzDRL} formalizes the agent-environment interface, introduces batched vectorization, and provides quantitative evaluation of state-transition and training reproducibility.

Game-engine-based simulators such as Flightmare \cite{song2021flightmare} and AirSim \cite{shah2017airsim} provide visually rich, high-speed environments using Unity and Unreal Engine, respectively. Recent GPU-accelerated frameworks such as Aerial Gym \cite{kulkarni2023aerial}, based on Isaac Gym, and OmniDrones \cite{xu2024omnidrones}, based on Omniverse Isaac Sim, provide massive parallelism using NVIDIA’s simulation stack. PyBullet-Drones \cite{panerati2021learning} offers synchronous CPU-based multirotor RL.

Unlike these prior works, \textit{GzDRL} enables scalable, deterministic RL experimentation directly within Gazebo, preserving full compatibility with built-in robot models, sensors, and ROS-based deployment workflows. By introducing direct ECS-based stepping, efficient vectorization, and zero-copy observation sharing, \textit{GzDRL} bridges the gap between high-throughput RL data collection and realistic robotics simulation, addressing key shortcomings of middleware and game-engine-based alternatives.

\section{\textit{GzDRL} Platform}
\textit{GzDRL} is a scalable reinforcement learning platform for robotics simulation, designed to unify high-fidelity Gazebo environments with efficient, reproducible RL experimentation. The platform integrates parameterizable multirotor models, a suite of control abstractions, and a single-process \textit{DRLServer} core that enables direct, deterministic agent-environment interaction without middleware. Two complementary vectorization pathways—Python-based and C++-based—leverage modern CPU architectures for high-throughput data collection. The execution details are provided in Sec.~\ref{sec:architecture}.
\subsection{Multi-rotor Dynamics \& Control}\label{sec:dynamics}
\textit{GzDRL} supports high-fidelity simulation of UAV rigid-body dynamics, following established models \cite{furrer2016rotors, hanover2024autonomous}. The UAV is treated as a six-degree-of-freedom (DoF) rigid body with mass $m$ and diagonal inertia matrix $\mathbf{J}$. The state vector $\mathbf{x}=\text{col}(\mathbf{p}_W, \mathbf{v}_W,\mathbf{q},\boldsymbol{\omega}_{B}) \in \mathbb{R}^{13}$ evolves according to:

\begin{equation}\label{eq:gen_dyn}
\begin{aligned}
\dot{\mathbf{x}} = f(\mathbf{x}, \mathbf{u}) = \begin{bmatrix}
\mathbf{v}_W\\
\frac{1}{m}(R \mathbf{T}_B + \mathbf{F}_W) + \mathbf{g}_W\\
\frac{1}{2}\mathbf{q} \otimes (0, \boldsymbol{\omega}_B)\\
\mathbf{J}^{-1}(\boldsymbol{\tau}_B + \mathbf{M}_B - \boldsymbol{\omega}_B \times \mathbf{J}\boldsymbol{\omega}_B)
\end{bmatrix}
\end{aligned}
\end{equation}
where $\mathbf{p}_W$ is the world-frame position, $\mathbf{v}_W$ is world-frame velocity, $\mathbf{q}$ and $R$ represent UAV orientation (quaternion and rotation matrix), $\boldsymbol{\omega}_{B}$ is the angular velocity in body frame, $\mathbf{T}_B = \text{col}(0,0, T)$ is collective thrust in the body frame, $\mathbf{F}_W$ and $\mathbf{M}_B$ are disturbance force/moment, $\mathbf{g}_W$ is gravity, $\boldsymbol{\tau}_B$ is net torque, and $\mathbf{u}=\text{col}(T, \boldsymbol{\tau}_B)$ is the control input.

\textbf{Control Abstractions:} \textit{GzDRL} implements a unified, modular interface supporting hierarchical control at multiple abstraction levels \cite{eschmann} (see Fig.~\ref{fig:control_heirs}):
\begin{figure}[htbp]
\centering
\includegraphics[width=0.8\columnwidth]{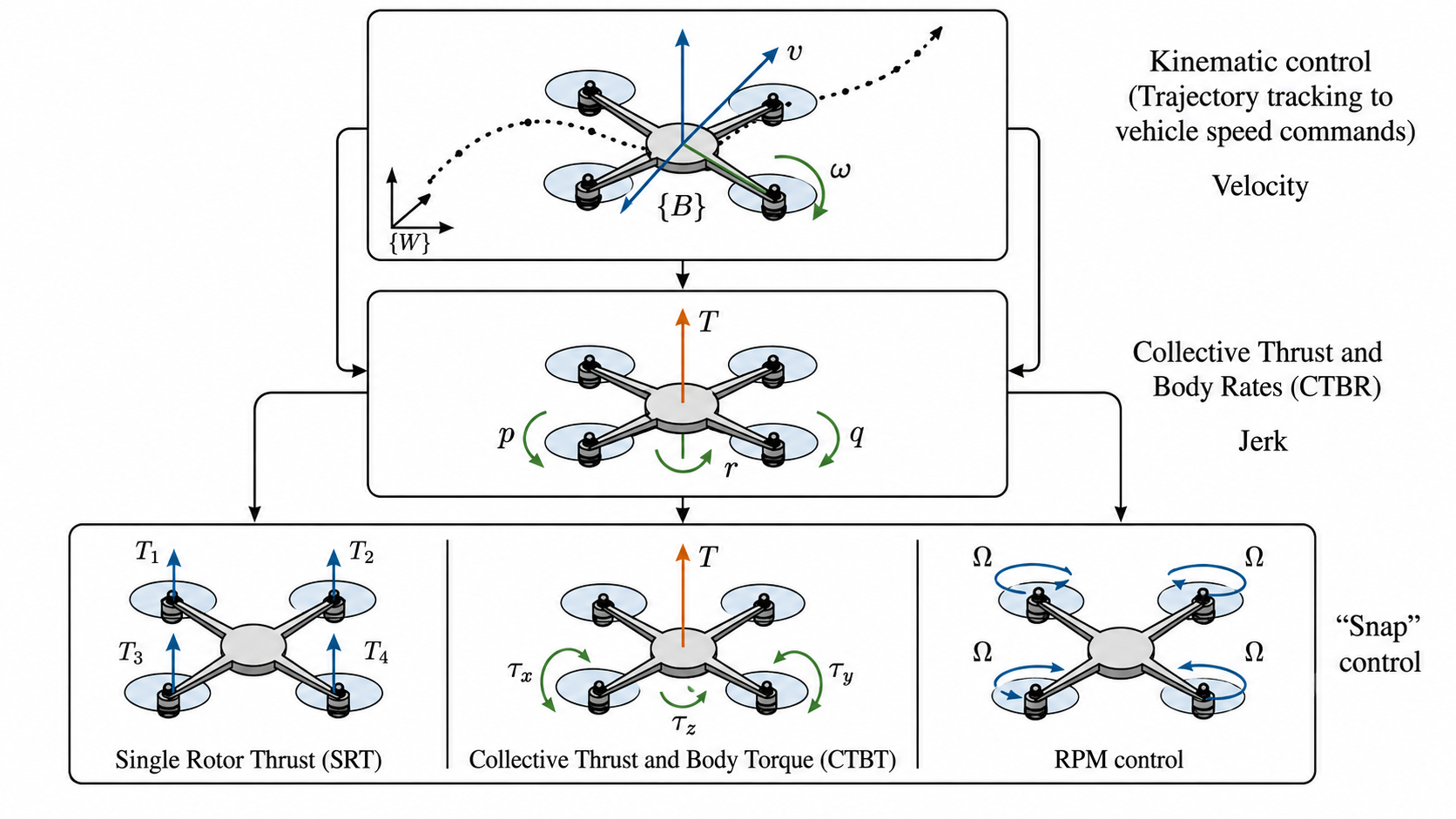}
\caption{Various control hierarchies supported by \textit{GzDRL}}
\label{fig:control_heirs}
\end{figure}
\begin{itemize}

\item \textbf{Velocity:} Kinematic position/velocity commands

\item \textbf{Jerk:} Collective thrust and body rate (CTBR)

\item \textbf{Snap:} Single rotor thrust (SRT), collective thrust/body torque (CTBT), and RPM commands
\end{itemize}This structure enables flexible prototyping and benchmarking across control modalities for RL research.

To model actuator and rotor lag, \textit{GzDRL} models first-order actuator dyanmics:

\begin{equation}\label{eq:rot_dyn}
\mathbf{u}(t+\Delta t) = \mathbf{u}(t)e^{-\Delta t/t_c} + \mathbf{u}_{\text{ref}}(t)(1-e^{-\Delta t/t_c})
\end{equation}
where $\Delta t$ is the sampling time and $t_c$ is the first-order time constant.

\textbf{Supported Control Modes:}

\begin{enumerate}

\item \textbf{RPM:} Policy outputs rotor velocities ($u \in \mathbb{R}^{N}$); forces/torques computed per \cite{furrer2016rotors}

\item \textbf{SRT:} Policy outputs single rotor thrust; yaw torques calculated from motor/geometry constants

\item \textbf{CTBT:} Policy specifies thrust and body torques ($u \in \mathbb{R}^4$) applied at the UAV CoG

\item \textbf{CTBR:} Policy specifies collective thrust and reference body rates; optional internal PD controller tracks rates

\item \textbf{Wrench:} Forces/torques in world frame ($u \in \mathbb{R}^6$) applied at chosen links

\item \textbf{Velocity/Angular Velocity:} Instantaneous kinematic commands

\item \textbf{High-Level Guidance:} Policy outputs state references for integrated controllers (e.g., geometric, NMPC)

\end{enumerate}

This comprehensive stack enables RL agents to interface at any abstraction level, supporting both end-to-end and modular policy architectures. Unified abstractions and high-fidelity dynamics make 
\textit{GzDRL} suitable for benchmarking, algorithm prototyping, and deployment-oriented RL research.
\begin{table*}[t]
\centering
\caption{
Comparison of multirotor simulation frameworks.
\yes{} denotes native/full support, \no{} no support, and
\partialyes{} support through an external bridge.
}
\label{tab:comparison}

\scriptsize
\setlength{\tabcolsep}{3.2pt}
\renewcommand{\arraystretch}{1.13}

\resizebox{\textwidth}{!}{%
\begin{tabular}{@{}l ll cc cc ll@{}}
\toprule

& \multicolumn{4}{c}{\textbf{Simulation}}
& \multicolumn{2}{c}{\textbf{Robotics Integration}}
& \multicolumn{2}{c}{\textbf{RL Interface}}
\\

\cmidrule(lr){2-5}
\cmidrule(lr){6-7}
\cmidrule(l){8-9}

\textbf{Framework}
&
\textbf{\makecell{Physics\\engine}}
&
\textbf{\makecell{Rendering\\engine}}
&
\textbf{\makecell{Runtime\\rand.}}
&
\textbf{\makecell{Sync.\\stepping}}
&
\textbf{\makecell{Native\\ROS}}
&
\textbf{Sensors}
&
\textbf{Language}
&
\textbf{\makecell{Agent\\support}}
\\

\midrule

RotorS \cite{furrer2016rotors}
& DART, Bullet, ODE
& OGRE 1.0
& \no
& \no
& \yes
& IMU, RGB-D
& C\texttt{++}
& \no
\\

AirSim \cite{shah2017airsim}
& PhysX
& Unreal
& \no
& \no
& \partialyes
& IMU, RGB-D, S
& C\texttt{++}, Py.
& Single, Multi
\\

Flightmare \cite{song2021flightmare}
& Flexible
& Unity
& \yes
& \yes
& \no
& IMU, RGB-D, S
& C\texttt{++}, Py.
& Single
\\

PyBullet-Drones \cite{panerati2021learning}
& Bullet
& OpenGL
& \no
& \yes
& \no
& IMU, RGB-D, S
& Python
& Single, Multi
\\

OmniDrones \cite{xu2024omnidrones}
& PhysX
& Omniverse RTX
& \yes
& \yes
& \partialyes
& IMU, RGB-D, S, F, C
& Python
& Single, Multi
\\

Aerial-Gym \cite{kulkarni2023aerial}
& PhysX
& Isaac Gym/Warp
& \yes
& \yes
& \no
& IMU, RGB-D, S, L
& Python
& Single, Multi
\\

\midrule

\rowcolor{gray!10}
\textbf{\textit{GzDRL} (ours)}
& \textbf{DART, Bullet, TPE}
& \textbf{OGRE 2.0, OptiX}
& \yes
& \yes
& \yes
& \textbf{IMU, RGB-D, S, F, C}
& \textbf{C\texttt{++}, Python}
& \textbf{Single, Multi}
\\

\bottomrule
\end{tabular}%
}

\vspace{2pt}
\begin{minipage}{0.99\textwidth}
\footnotesize
\textit{Sensor abbreviations:}
S-segmentation; F-force; C-contact; L-LiDAR.
``Flexible'' denotes support for externally coupled dynamics/rendering
components.
\end{minipage}

\end{table*}
\section{Execution Architecture}
\label{sec:architecture}

\textit{GzDRL}'s execution architecture is designed for reproducible, high-throughput RL data collection in Gazebo, eliminating middleware-related latency and nondeterminism. The system centers around a single-process \textit{DRLServer}, which interfaces directly with the Gazebo simulation core, and supports efficient vectorization for scalable experimentation. This section details the systematic environment-stepping mechanism, vectorization strategies, and design choices enabling deterministic agent-environment interaction, resource efficiency, and seamless integration with RL workflows. \subsection{Systematic Environment-stepping Mechanism }
Let $E_i$ denote a reinforcement learning environment managed by a \textit{DRLServer} containing a \texttt{gz::sim::Server} \cite {gazebo_architecture_docs}, and one Gazebo world. Gazebo API provides the \texttt{Server::Run} operation, which advances the simulation by one physics step. \textit{GzDRL} defines a systematic, sequential RL framework that is constructed around this operation. For a given policy action $\mathbf{a}_t$ and $K$ physics iterations, define $\mathbf{x}_{t,0}=\mathbf{x}_t$ as the initial environment state, \textit{GzDRL} executes the following procedure:
\begin{equation}
\begin{aligned}
\bar{\mathbf{a}}_t&=P_i(\mathbf{a}_t)\\
\mathbf{u}_{t,k}&=C_i(\bar{\mathbf{a}}_t,\mathbf{x}_{t,k-1})\quad k=1{:}K\\
\mathbf{u}_{t,k}&\xrightarrow{\mathrm{DRLServer}}\mathrm{Run}_i(1)
\xrightarrow{\mathrm{PostUpdate}}\mathbf{x}_{t,k}\\
(\mathbf{o}_{t+1},r_t,d_t)&=\Psi_i(\mathbf{x}_{t,K})
\end{aligned}
\label{eq:transition_contract}
\end{equation}
Here, $P_i$ processes the policy action once, $C_i$ represents an optional controller evaluated from the latest state, and $\Psi_i$ generates the RL observation, reward, and termination flag outputs.

Control commands are issued through the control abstraction interfaces in \textit{GzDRL} before simulation advancement. Each environment invokes \texttt{Server::Run} via \textit{DRLServer} $K$ times per policy action to progress the simulation. During each invocation, Gazebo executes its standard systems and physics updates. A custom \textit{DRLHelperSystem} reads poses, velocities,
accelerations, sensors, and contacts during \texttt{PostUpdate} \cite{gazebo_architecture_docs}, and performs model randomization or resets upon request. The complete environment state becomes accessible after the simulation has advanced.
Rendered sensors depend on Gazebo’s rendering pipeline and consume ECS state accordingly; the deterministic transition guarantees evaluated in Sec.~\ref{sec:synchronization_validation} apply specifically to state observations.

Each server is assigned a unique environment identifier, and construction is serialized to maintain consistency with the configuration of global static variables. Reset and domain randomization are managed by \textit{DRLServer} according to a request queue. Mass, inertia, and plugin parameters are modified in a cached SDF model, after which the
affected entity is respawned with the updated SDF, eliminating the need to reparse or reconstruct the world. Filter constants are updated directly.

\subsection{Vectorization}
\textit{AsyncDRLServerPool} provides the Python-based vectorization path for \textit{DRLServer}. It establishes a dedicated task queue and allocates a C\texttt{++} worker thread for each \textit{DRLServer}, utilizing CPU pinning on supported Linux systems. Python batches are dispatched in parallel and return after all executions in the batch are complete. By releasing the Python Global Interpreter Lock (GIL), \textit{AsyncDRLServerPool} enables high-speed execution of RL environments in Python.

For C+±based vectorization, \textit{GazeboPool} extends the EnvPool core \cite{weng2022envpool}. It maintains $N_E$ environments, $N_T$ threads, an action queue, and rotating C\texttt{++}-owned state
buffers. Threads dynamically dequeue environment-indexed requests and are not statically assigned to a specific \textit{DRLServer}. A slot in the state-buffer queue is committed after receiving the observation,
reward, and termination flags, and Python receives a zero-copy
NumPy view of the final block. In batched execution mode, a \texttt{recv} \cite{weng2022envpool} call returns after $N_B\leq N_E$ environment steps. In multi-robot tasks, a single $E_i$ contains
multiple interacting models, with commands applied before simulation advancement. A joint observation, team reward, and shared termination flags are then returned as the observation.

Algorithm~\ref{alg:gazebopool} summarizes the learner and worker sides of this execution flow.
\begin{algorithm}[htbp]
\caption{GazeboPool execution}
\label{alg:gazebopool}
\footnotesize
\begin{algorithmic}[1]
\REQUIRE $N_E$ environments, $N_T$ threads, batch-size $N_B$, $K$ physics-steps per action
\STATE Construct persistent $E_i=(\mathrm{task}_i,\mathrm{DRLServer}_i,
\mathrm{Gazebo}_i)$
\STATE Start workers, action queue $Q_A$, and state buffers $Q_S$
\STATE Submit one reset request for every $E_i$
\STATE \textbf{Learner loop:}
\LOOP
\STATE $B\gets Q_S.\operatorname{wait}(N_B)$; $A\gets\pi(B.\mathrm{obs})$
\FORALL{$(i,\mathbf{a}_i)\in(B.\mathrm{env\_id},A)$}
\STATE Store $\mathbf{a}_i$ and push $i$ to $Q_A$
\ENDFOR
\ENDLOOP
\STATE \textbf{Each worker executes concurrently:}
\LOOP
\STATE $i\gets Q_A.\operatorname{pop}()$
\IF{$E_i$ is terminal or explicitly reset}
\STATE Reset $E_i$ and construct its initial observation
\ELSE
\STATE $\bar{\mathbf{a}}\gets E_i.\operatorname{ProcessAction}(\mathbf{a}i)$
\FOR{$k=1$ to $K$}
\STATE Read latest environment state $\mathbf{x}_{k-1}$
\STATE $\mathbf{u}_k\gets E_i.\operatorname{Control}(\bar{\mathbf{a}},\mathbf{x}_{k-1})$
\STATE Write all $\mathbf{u}_k$; $\mathrm{Server}_i.\operatorname{Run}(1)$
\ENDFOR
\STATE Read final \texttt{PostUpdate} state; compute $(\mathbf{o},r,d)$
\ENDIF
\STATE Write $(i,\mathbf{o},r,d)$ to a free slot and commit it
\ENDLOOP
\end{algorithmic}
\end{algorithm}
\section{Experiments}
We rigorously evaluate \textit{GzDRL} with respect to reproducibility, scalability, technical rigor, and real-world relevance, benchmarking against established baselines. Experiments were conducted on both a workstation (AMD Ryzen Threadripper 3960X, 256 GB RAM, dual NVIDIA RTX A6000 GPUs) and a laptop (AMD Ryzen 9 7940HS, 32 GB RAM, NVIDIA RTX 4060). Throughput is reported as the number of completed environment transitions per unit wall time. All experiments were run in headless mode for fair comparison.

We evaluate \textit{GzDRL} in terms of state-transition and training reproducibility, scalability across single- and multi-agent settings, policy learning with various control abstractions, domain randomization capabilities, and direct physical deployment.


\subsection{State Transition Synchronization and Reproducibility Validation}
\label{sec:synchronization_validation}
\begin{figure}[htbp]
    \centering
    \includegraphics[width=\columnwidth]{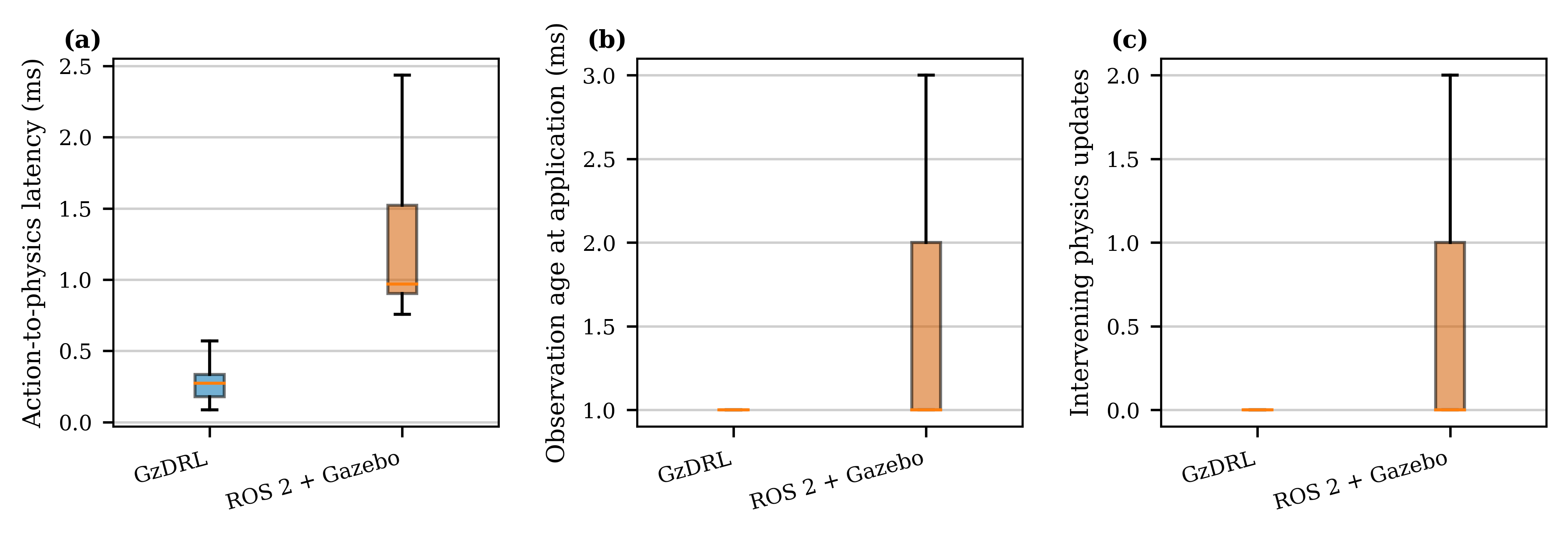}
    \caption{
Transition synchronization metrics over 100 trials: (a) Latency between action and physics update; (b) Observation delay; (c) Intervening physics steps. \textit{GzDRL} achieves deterministic, minimal-latency transitions, while ROS~2 incurs variable delays.
}
\label{fig:synchronization_timing}
\end{figure}

We benchmarked \textit{GzDRL} against ROS~2-Gazebo to assess temporal consistency and reproducibility in agent-environment interactions, using identical tasks, timing, and initial states. The only difference was that \textit{GzDRL} used synchronous single-process stepping, while ROS~2 relied on asynchronous topic callbacks.
Synchronization was quantified via three metrics: (i) latency between action and physics update, (ii) observation delay, and (iii) intervening physics steps between observation and action application. As shown in Fig.~\ref{fig:synchronization_timing}, \textit{GzDRL} consistently applies policy actions at the immediately succeeding physics update, maintaining minimal latency. In contrast, ROS~2 suffers from variable multi-step delays due to middleware.

\begin{figure}[htbp]
    \centering
    \includegraphics[width=\columnwidth]{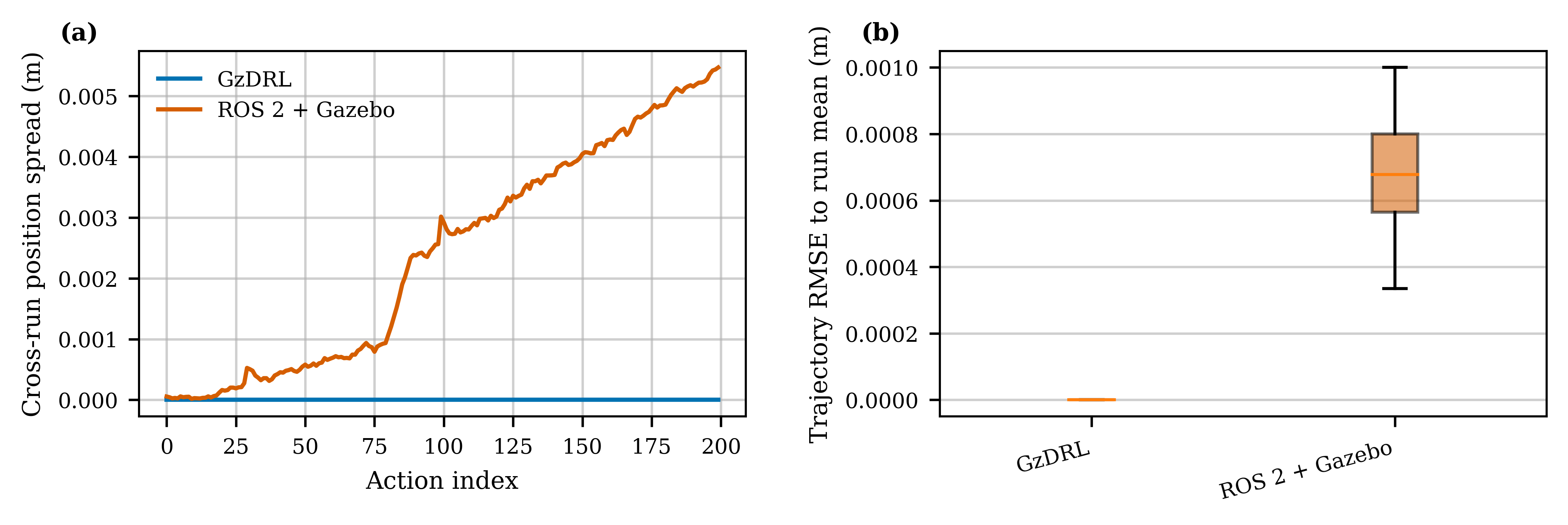}
    \caption{
Cross-run trajectory reproducibility: (a) Position spread versus action index; (b) Trajectory RMSE distribution over 100 trials. \textit{GzDRL} maintains coincident trajectories; ROS~2 introduces cumulative dispersion.
}
\label{fig:trajectory_reproducibility}
\end{figure}
Trajectory reproducibility results (Fig.~\ref{fig:trajectory_reproducibility}) further distinguish the platforms: \textit{GzDRL} yields effectively identical trajectories across 100 trials, while ROS~2 accumulates measurable divergence. These findings confirm that \textit{GzDRL}’s architecture ensures deterministic, reproducible RL data, directly overcoming a key bottleneck in scalable, robust robotics learning.

\subsection{Training Reproducibility}
\label{sec:training_reproducibility}
We assessed whether the deterministic state-transition feature of \textit{GzDRL} yields reproducible reinforcement learning experiments. Using the \textit{GazeboPool} hover task, we executed two groups of five PPO \cite{schulman2017proximal} training runs in separate processes: one with identical RL seeds (\emph{same-seed}) and one with distinct seeds (\emph{different-seed}: $\{1,5,25,42,125\}$). All other parameters—including environment seed, algorithm configuration, normalization, and hardware—were held fixed, and domain randomization was disabled.

Each policy was trained for $5\times10^5$ environment steps with eight parallel environments, saving checkpoints every $5\times10^4$ steps. Every checkpoint was evaluated deterministically over 20 fixed-seed episodes in a new process, with observation normalization restored and frozen. 
\begin{figure}[htbp]
\centering
\includegraphics[width=\columnwidth]{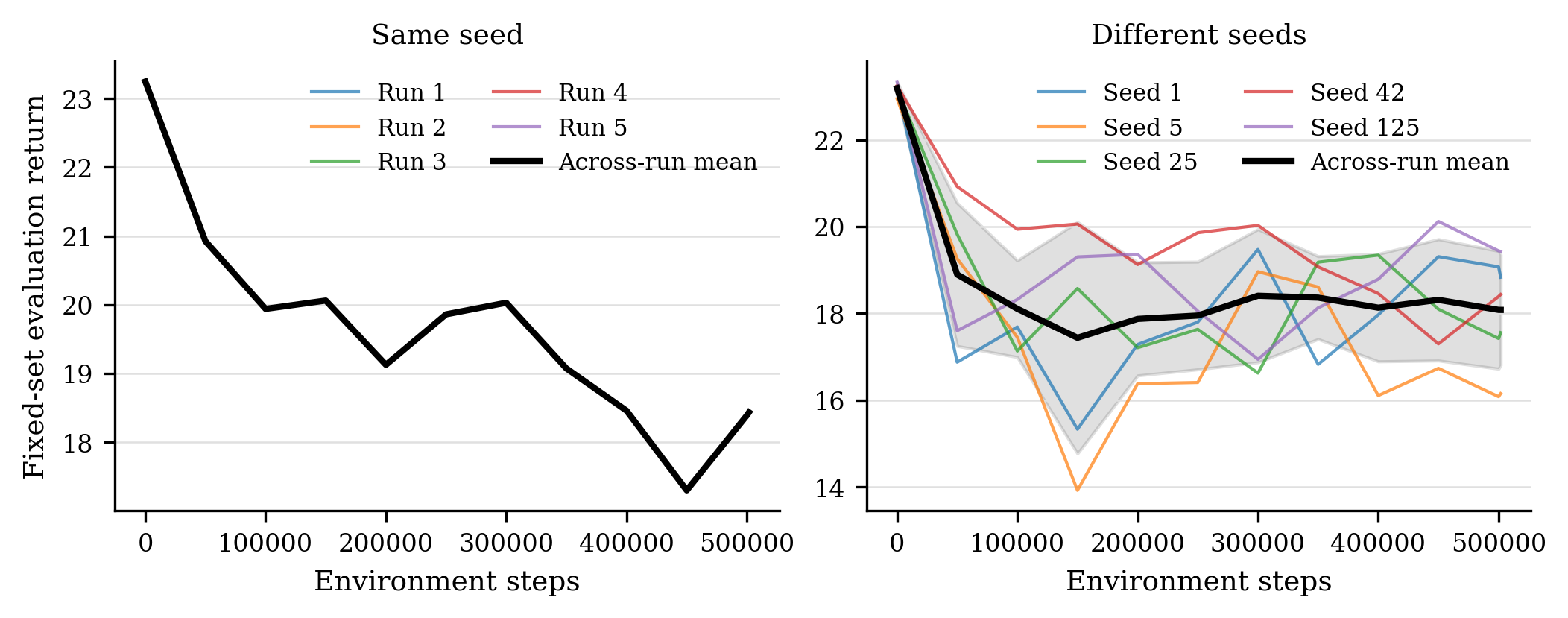}
\caption{Training reproducibility on the \textit{GzDRL} hover task. Five independent runs with the same RL seed (left) produce coincident learning curves, while five different seeds (right) yield substantial dispersion. Each checkpoint is evaluated on the same fixed episode set; shaded region denotes $\pm 1$ standard deviation.}
\label{fig:training_reproducibility}
\end{figure}
\begin{table}[htbp]
\centering
\caption{Training reproducibility on the GzDRL hover task over five
process-isolated runs per configuration. Final return is reported as
mean $\pm$ standard deviation across runs; all checkpoints are evaluated
on the same fixed deterministic episode set.}
\label{tab:training_reproducibility}
\footnotesize
\setlength{\tabcolsep}{4pt}
\renewcommand{\arraystretch}{1.12}

\begin{tabular}{lrrrr}
\toprule
\textbf{Configuration}
& \textbf{Final return}
& \textbf{CV (\%)}
& \makecell{\textbf{Mean ckpt.}\\$\boldsymbol{\sigma}$}
& \makecell{\textbf{Identical}\\\textbf{ckpts. (\%)}} \\
\midrule
Same seed
& $18.43 \pm 0.00$
& 0.00
& 0.00
& 100 \\
Different seeds
& $18.08 \pm 1.27$
& 7.05
& 1.32
& 0 \\
\bottomrule
\end{tabular}
\end{table}

As shown in Fig.~\ref{fig:training_reproducibility} and Table~\ref{tab:training_reproducibility}, the five runs that use the same seed produce identical evaluation curves. The standard deviation of the final return is zero, and the mean standard deviation of the return across checkpoints is also zero.
Furthermore, 100\% of the corresponding policy checkpoints have identical parameter hashes. In contrast, the runs that use different seeds show substantial variation. The coefficient of
variation of the final return is 7.05\%, and the mean standard deviation across checkpoints is 1.32. These results demonstrate that \textit{GzDRL} supports reproducible RL training.
\subsection{Scalability and Middleware Overhead Validation}
\label{sec:scalability}
We evaluated the computational scalability of \textit{GzDRL} using a standardized hover task with identical observation and action spaces, task logic, and a fixed physics timestep ($\Delta t=1$~ms) across all simulators. The number of parallel environments was varied as $N_E\in\{1,4,8,16,32,64,128\}$. Each configuration was run for five independent trials, with a 1,000-step warmup and 10,000 measured environment steps. We report environment throughput and CPU/GPU utilization on both workstation and laptop platforms.

Baselines included PyBullet-Drones~\cite{panerati2021learning}, OmniDrones~\cite{xu2024omnidrones}, Aerial Gym~\cite{kulkarni2023aerial}, and conventional ROS~2–Gazebo with serial and process-based vectorization. Within \textit{GzDRL}, we compared \textit{GazeboPool}, \textit{AsyncDRLServerPool}, and Python-based parallelization schemes. For \textit{GazeboPool}, batch and worker counts were set to $N_B \approx N_E/3$, and $N_T=\min(N_E,N_{\text{CPU}})$.
\begin{figure}[htbp]
\centering
\includegraphics[width=\columnwidth]{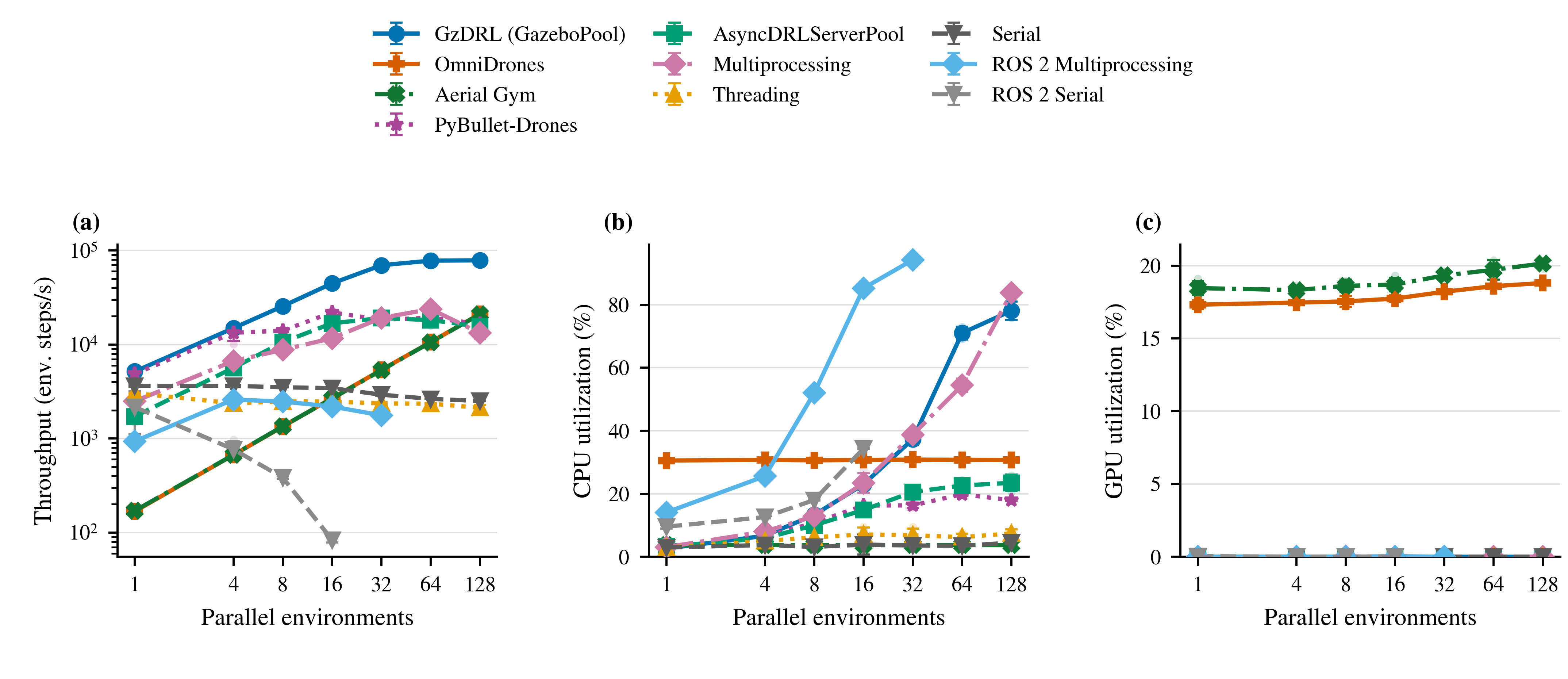}
\includegraphics[width=\columnwidth]{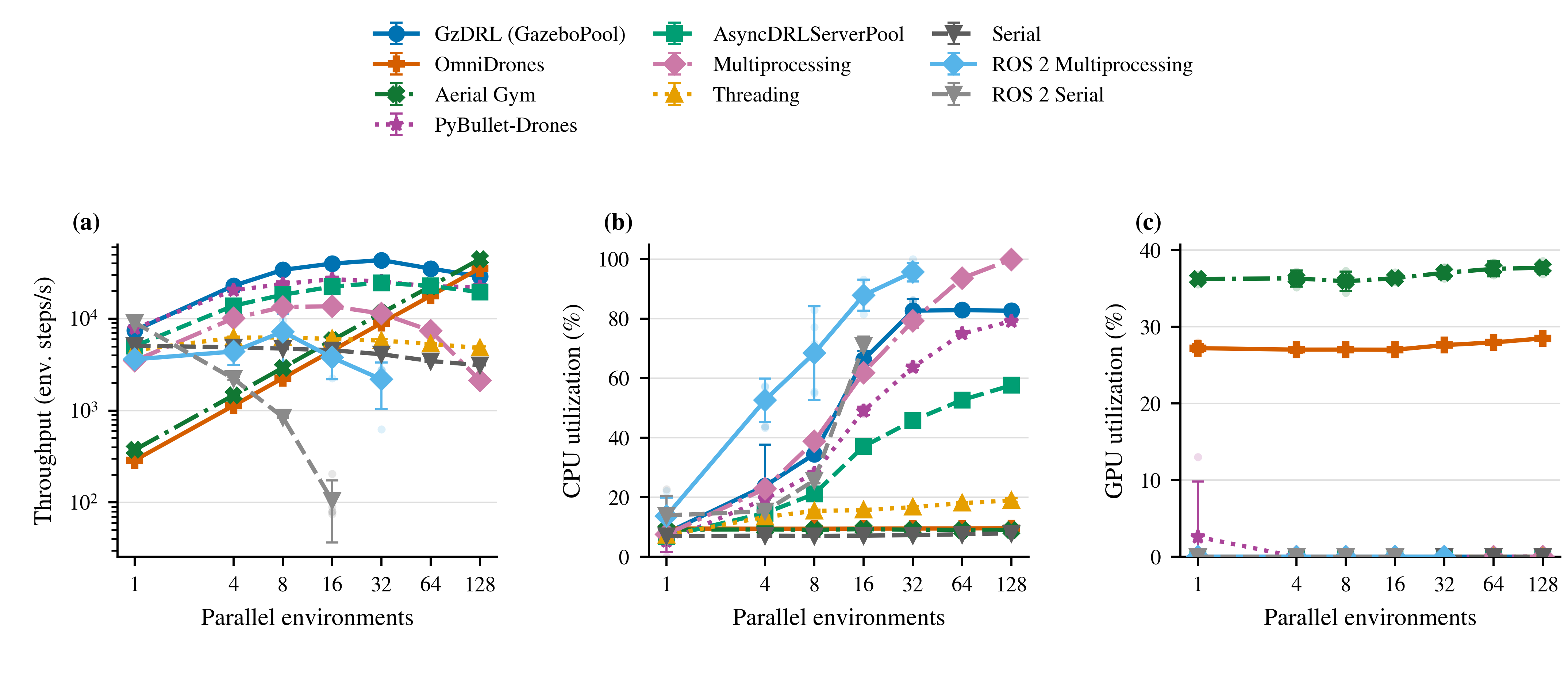}
\caption{Backend scalability on workstation (top) and laptop (bottom): (a) environment throughput, (b) CPU utilization, (c) GPU utilization. Markers indicate trial means; error bars show 95\% confidence intervals.}
\label{fig:ipc_scalability}
\end{figure}

Figure~\ref{fig:ipc_scalability} and Table~\ref{tab:backend-peak-comparison} summarize the results of the scalability experiment. Figure~\ref{fig:ipc_scalability} shows that \textit{GazeboPool} achieves monotonic scaling with increasing $N_E$ until CPU saturation. On the workstation, \textit{GzDRL} reaches $78.6{\times}10^3$ steps/s at $N_E{=}128$, outperforming PyBullet-Drones ($22.0{\times}10^3$), OmniDrones ($21.2{\times}10^3$), and Aerial Gym ($21.0{\times}10^3$), a $\sim$3.6$\times$ advantage over the fastest baseline. On the laptop, \textit{GzDRL} peaks at $43.4{\times}10^3$ steps/s ($N_E{=}32$), remaining competitive with GPU-accelerated Aerial Gym ($44.6{\times}10^3$), and exceeding OmniDrones and PyBullet-Drones.
Conventional ROS~2–Gazebo vectorization scaled poorly, with peak workstation throughput at $2.6{\times}10^3$ steps/s (process-based), giving \textit{GzDRL} a $30{\times}$ throughput advantage. Internal baselines demonstrated that threading is constrained by Python/GIL overhead, serial provides no parallelism, and multiprocessing incurs high overhead; among the evaluated \textit{GzDRL} variants, \textit{GazeboPool} achieves the strongest scalability and highest throughput.
These results demonstrate that bypassing middleware and leveraging batched vectorization yields both deterministic state transitions and the highest measured throughput, validating the \textit{GzDRL} architecture for scalable robotics RL.
\begin{table}[t]
\centering
\caption{Backend scalability. Peak rate is the maximum mean environment throughput over the tested $N_E$ values and is reported as mean $\mathbin{\pm}$ 95\% Student's $t$ confidence interval. $N_E^\star$, CPU, and GPU utilization are evaluated at the same operating point. Rates are in $10^3$ environment steps/s.}
\label{tab:backend-peak-comparison}
\scriptsize
\setlength{\tabcolsep}{3.0pt}
\renewcommand{\arraystretch}{1.08}
\resizebox{\columnwidth}{!}{%
\begin{tabular}{lrrrrrrrr}
\toprule
 & \multicolumn{4}{c}{\textbf{Workstation}} & \multicolumn{4}{c}{\textbf{Laptop}} \\
\cmidrule(lr){2-5} \cmidrule(lr){6-9}
\textbf{Architecture} & \textbf{Peak rate} & \textbf{$N_E^\star$} & \textbf{CPU (\%)} & \textbf{GPU (\%)} & \textbf{Peak rate} & \textbf{$N_E^\star$} & \textbf{CPU (\%)} & \textbf{GPU (\%)} \\
\midrule
GzDRL (GazeboPool) & \textbf{\(78.58 \mathbin{\pm} 2.08\)} & 128 & 78.1 & 0.0 & \(43.38 \mathbin{\pm} 0.56\) & 32 & 82.7 & 0.0 \\
OmniDrones & \(21.15 \mathbin{\pm} 0.22\) & 128 & 30.7 & 18.8 & \(35.47 \mathbin{\pm} 0.20\) & 128 & 9.6 & 28.5 \\
Aerial Gym & \(21.02 \mathbin{\pm} 0.27\) & 128 & 3.7 & 20.1 & \textbf{\(44.56 \mathbin{\pm} 0.42\)} & 128 & 9.0 & 37.7 \\
PyBullet-Drones & \(21.95 \mathbin{\pm} 1.25\) & 16 & 16.3 & 0.0 & \(26.92 \mathbin{\pm} 1.30\) & 16 & 49.0 & 0.0 \\
GzDRL (AsyncDRLServerPool) & \(19.10 \mathbin{\pm} 1.79\) & 32 & 20.6 & 0.0 & \(24.59 \mathbin{\pm} 1.37\) & 32 & 45.8 & 0.0 \\
GzDRL (Multiprocessing) & \(23.80 \mathbin{\pm} 0.42\) & 64 & 54.5 & 0.0 & \(13.64 \mathbin{\pm} 0.66\) & 16 & 61.9 & 0.0 \\
GzDRL (Threading) & \(3.03 \mathbin{\pm} 0.03\) & 1 & 3.0 & 0.0 & \(6.24 \mathbin{\pm} 0.13\) & 8 & 15.4 & 0.0 \\
GzDRL (Serial) & \(3.63 \mathbin{\pm} 0.06\) & 1 & 3.0 & 0.0 & \(5.08 \mathbin{\pm} 0.12\) & 1 & 7.0 & 0.0 \\
ROS 2 Multiprocessing & \(2.59 \mathbin{\pm} 0.18\) & 4 & 25.6 & 0.0 & \(7.22 \mathbin{\pm} 4.01\) & 8 & 68.4 & 0.0 \\
ROS 2 Serial & \(2.16 \mathbin{\pm} 1.05\) & 1 & 9.6 & 0.0 & \(9.06 \mathbin{\pm} 0.81\) & 1 & 13.9 & 0.0 \\
\bottomrule
\end{tabular}%
}
\end{table}

\begin{table}[t]
\centering
\caption{Multi-agent throughput. Entries are trial-mean environment throughput in $10^3$ steps/s; uncertainty is reported in the corresponding scalability figure.}
\label{tab:multiagent-throughput}
\scriptsize
\setlength{\tabcolsep}{3.5pt}
\renewcommand{\arraystretch}{1.08}
\resizebox{\columnwidth}{!}{%
\begin{tabular}{llrrrr}
\toprule
\textbf{Platform} & \textbf{Architecture} & \textbf{$N_A=2$} & \textbf{$N_A=5$} & \textbf{$N_A=10$} & \textbf{$N_A=20$} \\
\midrule
Workstation & GzDRL (GazeboPool) & 35.52 & 17.82 & 8.10 & 3.18 \\
 & OmniDrones & 4.24 & 4.19 & 4.15 & 4.10 \\
 & Aerial Gym & 7.15 & 7.07 & 6.62 & 5.53 \\
 & PyBullet-Drones & 13.89 & 10.23 & 6.65 & 2.69 \\
\addlinespace[1pt]
Laptop & GzDRL (GazeboPool) & 32.95 & 11.54 & 5.01 & 2.16 \\
 & OmniDrones & 7.11 & 6.94 & 6.89 & 6.77 \\
 & Aerial Gym & 15.24 & 14.88 & 14.04 & 11.44 \\
 & PyBullet-Drones & 13.31 & 9.07 & 5.09 & 2.11 \\
\bottomrule
\end{tabular}%
}
\end{table}

GPU-based frameworks, including Aerial Gym and OmniDrones, become increasingly competitive as per-environment workload increases, as demonstrated by the multi-agent experiments in Sec.~\ref{sec:multiagent_scalability}. Nonetheless, \textit{GzDRL} achieves the highest measured workstation throughput among the evaluated frameworks, delivering high scalability and accessibility without reliance on GPU acceleration.
\subsection{Multi-Agent Scalability}
\label{sec:multiagent_scalability}
We evaluated simulator throughput as a function of the number of interacting UAVs per environment, using $N_A\in\{2,5,10,20\}$ agents and maintaining 32 parallel environments. \textit{GzDRL} was compared against OmniDrones, Aerial Gym, and PyBullet-Drones, all operating with a 1~ms physics timestep, and with matched task logic, robot count, and action/observation spaces. Each configuration was assessed over five independent trials with a 1,000-step warmup and 10,000 measured steps; CPU and GPU utilization were recorded alongside throughput.
\begin{figure}[htbp]
\centering
\includegraphics[width=\columnwidth]{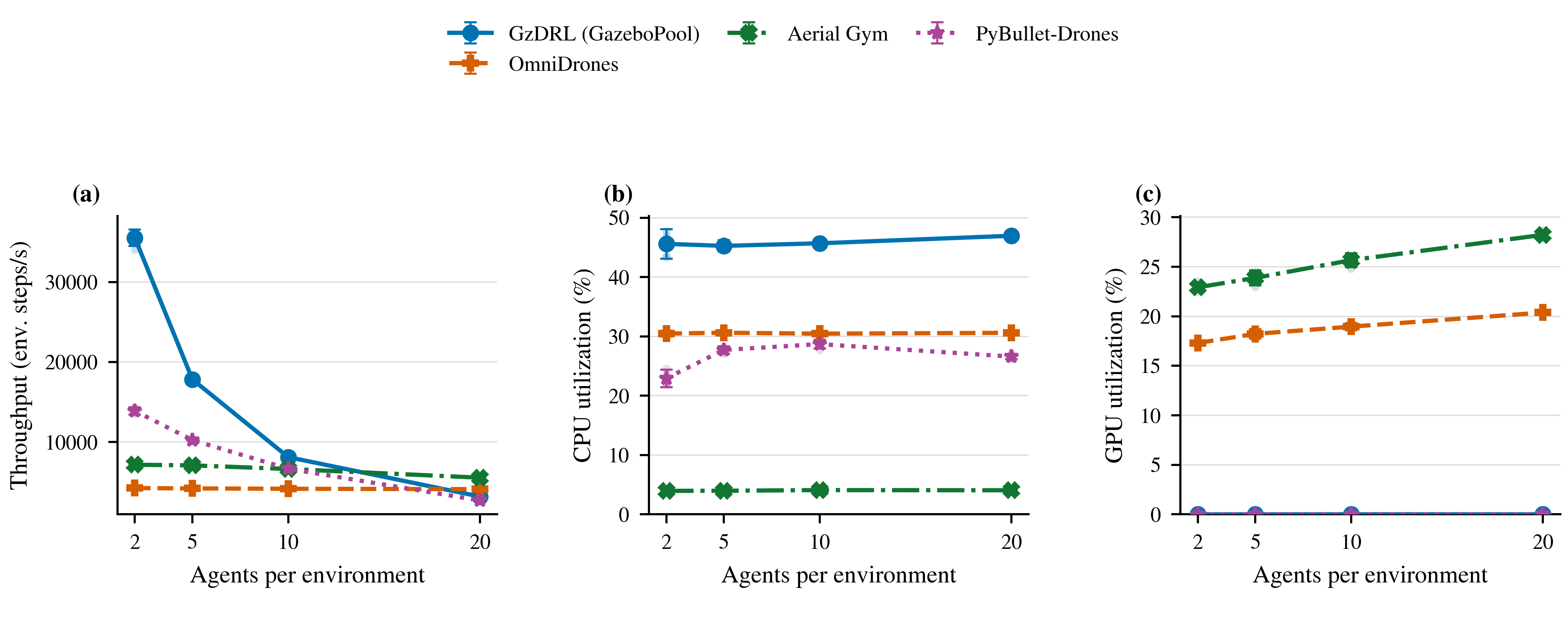}
\includegraphics[width=\columnwidth]{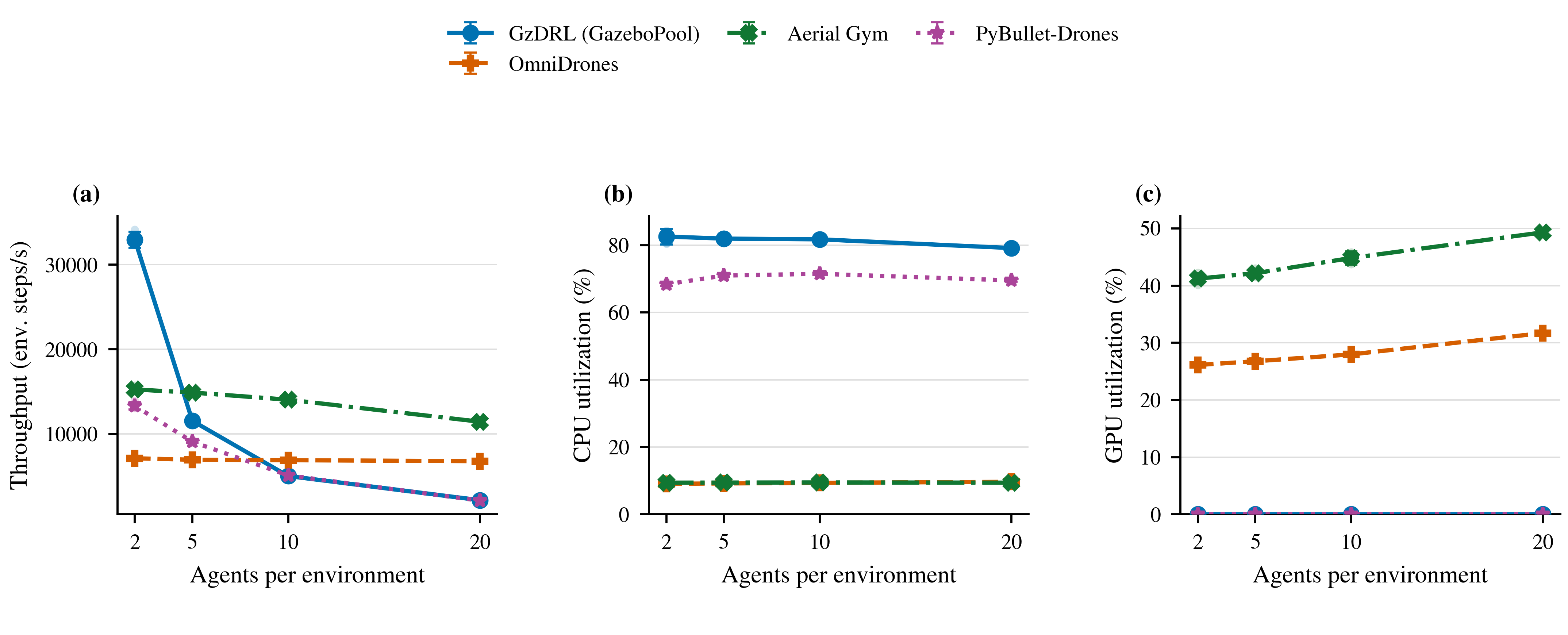}
\caption{Multi-agent scalability on workstation (top) and laptop (bottom): (a) environment throughput, (b) CPU utilization, (c) GPU utilization as the number of interacting UAVs per environment increases.}
\label{fig:vectorization}
\end{figure}

Figure~\ref{fig:vectorization} and Table~\ref{tab:multiagent-throughput} summarize the results. On the workstation, \textit{GzDRL} achieves $35.5\times10^3$, $17.8\times10^3$, and $8.1\times10^3$ steps/s for $N_A=2$, $5$, and $10$, respectively, outperforming all baselines at these agent counts ($2.6\times$ faster than PyBullet-Drones at $N_A=2$). For larger agent counts ($N_A=20$), Aerial Gym attains $5.5\times10^3$ steps/s versus $3.2\times10^3$ for \textit{GzDRL}, illustrating the growing advantage of GPU-parallelism as task complexity increases.

On the laptop, \textit{GzDRL} maintains the highest throughput at $N_A=2$ ($32.9\times10^3$ steps/s), but Aerial Gym leads for $N_A\geq5$, reflecting the greater efficiency of GPU-based simulation for large multi-agent workloads. Notably, no fixed crossover point is observed; the transition between CPU and GPU performance dominance depends jointly on agent count and available hardware.

\subsection{Policy Training and Control Abstractions}
\begin{figure*}[htbp]
\centering
\includegraphics[width=\textwidth]{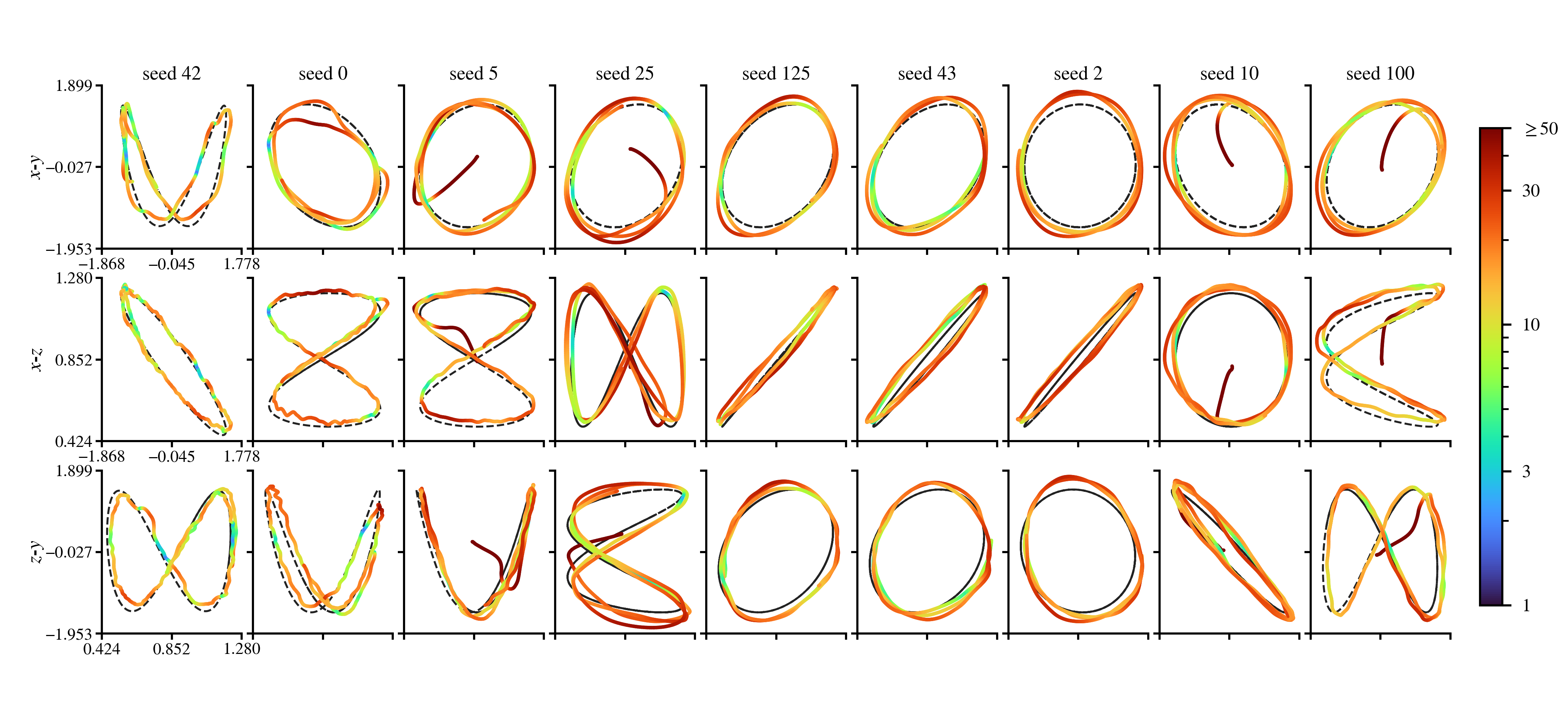}
\caption{Hardware trajectory tracking for nine reference trials. Each column: one flight; rows: three orthogonal projections. Dashed: commanded reference; solid: measured trajectory; color: instantaneous position error. The same frozen policy is used for all trials.}
\label{fig:hardware-trajectories}
\end{figure*}
To validate \textit{GzDRL}'s flexibility and RL relevance, we trained Proximal Policy Optimization (PPO) \cite{schulman2017proximal} policies using Stable-Baselines3~\cite{raffin2021stable} for two representative tasks: trajectory tracking and payload transport. Both end-to-end (rotor speed) and modular (3D force vector to an integrated attitude controller) control abstractions were benchmarked.

Each configuration used a [512,512] MLP, three random seeds ({1, 25, 42}), 15 parallel environments, and $5\times10^6$ transitions. Action and observation dimensions matched the control interface: (4,~700) (end-to-end) and (3,~680) (modular) for trajectory tracking at 1~kHz physics/100~Hz control, and (4,~26) and (3,~25) for payload transport at 1~kHz. All scenarios in a task were trained concurrently, yielding a total of 90 parallel environments (3 seeds $\times$ 15 environments per scenario $\times$ 2 control abstractions), and completed within 220 minutes of wall-clock time. Policy steps executed $K = f_{\mathrm{physics}}/f_{\mathrm{control}}$ physics iterations.
\begin{figure}[htbp]
\centering
\includegraphics[width=\columnwidth]{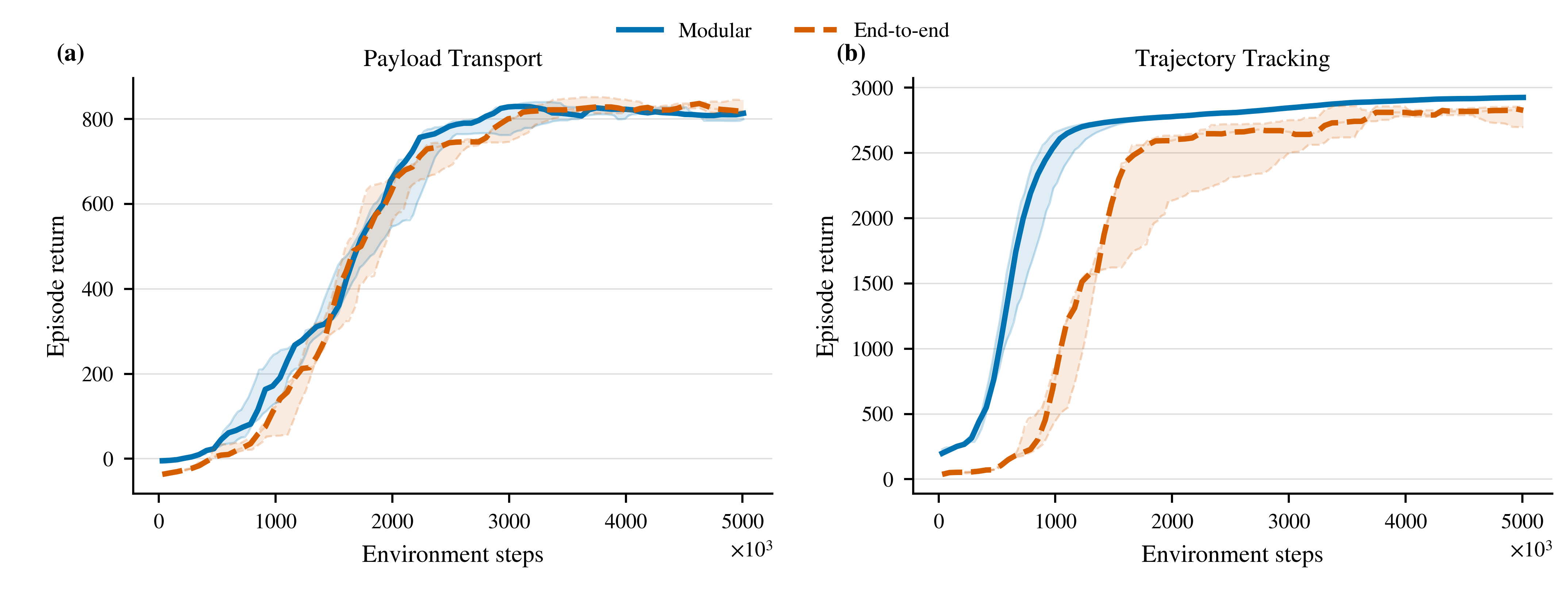}
\caption{Learning curves for payload-transport (left) and trajectory-tracking (right) tasks.}
\label{fig:training_curves}
\end{figure}
All configurations learned the corresponding tasks (Fig.~\ref{fig:training_curves}). Notably, hierarchical (modular) control accelerated early learning for the trajectory task, supporting the utility of abstraction layers in RL for robotics. These results demonstrate that \textit{GzDRL} enables rapid prototyping and benchmarking of both end-to-end and modular RL architectures, directly addressing experimental flexibility and deployment relevance.
\subsection{Domain Randomization}

To validate model randomization as a tool for sim-to-real transfer, we trained policies for the inverted pendulum on quadrotor task under two regimes: fixed model parameters and periodic parameter randomization every 10 episodes. Mass $m$, inertia $J$, and actuator time constants $(t_{\mathrm{up}}, t_{\mathrm{down}})$ were independently scaled by $k \in [0.9, 1.1]$ around nominal values, with all changes applied via \textit{DRLServer} during resets.
\begin{figure}[htbp]
\centering
\includegraphics[width=\columnwidth,trim={0 0 0 0},clip]{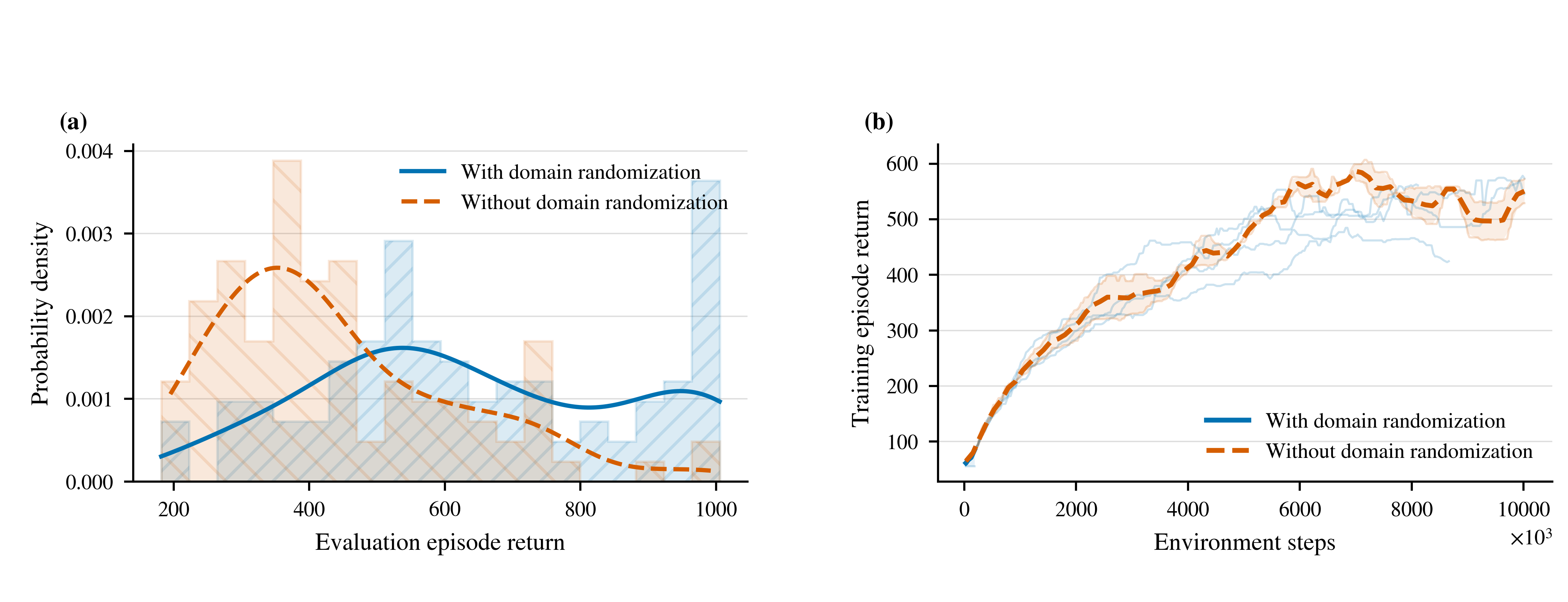}
\caption{Inverted-pendulum training (left) and evaluation under randomized
dynamics (right).}
\label{fig:domain_randomization}
\end{figure}
Figure~\ref{fig:domain_randomization} summarizes training and evaluation outcomes. When evaluated on 1,000 randomized episodes, the fixed-parameter policy achieved a reward of $439.89\pm175.97$, while the randomized policy achieved $642.97\pm232.07$. The randomized agent learned more conservatively ($398.45\pm130.33$ vs. $427.98\pm138.62$ during training), but generalized substantially better at test time. These results demonstrate runtime model randomization and improved policy generalization under parameter variation.

\subsection{Hardware Transfer and Onboard Deployment}
\label{sec:hardware}
We evaluated sim-to-real transfer by deploying a frozen trajectory-tracking policy, trained entirely in simulation, on a Quanser QDrone2 without additional fine-tuning. Onboard inference was performed at 100~Hz on an NVIDIA Jetson Xavier NX, with policy outputs tracked by the vehicle’s inner-loop controller and position feedback from a 120~Hz OptiTrack motion capture system.
Nine independent hardware trials were conducted, each following a distinct, seeded reference trajectory spanning diverse geometries and dynamic conditions. Performance metrics were computed over the moving-reference portion of each trial, excluding takeoff and terminal intervals.
\begin{table}[htbp]
\centering
\caption{Sim-to-real trajectory tracking on QDrone2 across nine reference trajectories. Values are mean $\pm$ 95\% confidence intervals.}
\label{tab:hardware-transfer}
\scriptsize
\setlength{\tabcolsep}{3.2pt}
\renewcommand{\arraystretch}{1.08}
\begin{tabular}{lc}
\toprule
\textbf{Metric} & \textbf{Hardware result} \\
\midrule
Mean Euclidean error [cm]              & $20.2 \pm 3.4$ \\
3D RMS error [cm]                      & $26.2 \pm 7.4$ \\
$\mathrm{RMSE}_x$ [cm]                 & $15.7 \pm 6.4$ \\
$\mathrm{RMSE}_y$ [cm]                 & $20.0 \pm 5.9$ \\
$\mathrm{RMSE}_z$ [cm]                 & $2.9 \pm 0.4$ \\
\bottomrule
\end{tabular}
\end{table}
The deployed policy achieved a mean position error of $20.2\pm3.4$~cm and a 3D RMS error of $26.2\pm7.4$~cm across trials, with consistent vertical tracking ($2.9\pm0.4$~cm RMSE). The 95th-percentile position error averaged $50.4\pm19.2$~cm. Peak reference speeds and accelerations reached $0.72\pm0.14$~m/s and $0.37\pm0.12$~m/s$^2$, demonstrating tracking across varying trajectories. No safety events or actuator saturations were observed.

Figure~\ref{fig:hardware-trajectories} visualizes reference and measured trajectories for all trials, with the frozen policy consistently following commanded paths despite variation in geometry and speed. All flights remained within actuation limits and the $20^\circ$ tilt safety threshold.
These results demonstrate that policies trained with \textit{GzDRL} can be deployed onboard a physical multirotor without adaptation, achieving stable and accurate trajectory tracking across diverse real-world scenarios. This establishes the deployment relevance and practical impact of the proposed framework beyond simulation benchmarks.

\section{Conclusion}
We introduced \textit{GzDRL}, a reinforcement learning framework that enables scalable, deterministic environment stepping and high-throughput data collection within Gazebo. By eliminating middleware-related nondeterminism, \textit{GzDRL} supports reproducible RL experimentation through direct, single-process control and efficient vectorization. Experiments show that \textit{GzDRL} achieves the highest measured single-agent throughput on the workstation CPU among the evaluated frameworks and remains competitive with GPU-based simulators on a laptop. In multi-agent settings, \textit{GzDRL} leads for small-to-moderate UAV counts on a workstation, while GPU simulators excel as the agent count grows. The framework also supports domain randomization and direct hardware transfer, demonstrating practical utility for sim-to-real robotics research. By addressing core bottlenecks in robotics RL, \textit{GzDRL} establishes a reproducible and scalable platform for high-fidelity learning and deployment.

\bibliographystyle{IEEEtran}
\bibliography{ref}

\end{document}